\documentclass[journal]{IEEEtran}
\usepackage{multirow}
\usepackage{subfigure}
\usepackage{graphicx}
\usepackage{amsmath}
\usepackage{amsthm}
\usepackage{booktabs}
\usepackage{algorithm}
\usepackage{algpseudocode}
\usepackage{xcolor}

\usepackage{changes}
\definechangesauthor[name={feng}, color=orange]{feng}
\usepackage{enumitem}

\ifCLASSINFOpdf

\else
  
\fi

\begin{document}

\title{Predict-then-Decide: A Predictive Approach for Wait or Answer Task in Dialogue Systems}

\author{Zehao~Lin, ~\IEEEmembership{Student Member, IEEE}, 
        Shaobo~Cui,
        Guodun~Li, Xiaoming~Kang, Feng~Ji, Fenglin~Li, Zhongzhou~Zhao, Haiqing~Chen, and Yin~Zhang* \thanks{* Corresponding Author}, ~\IEEEmembership{Member, IEEE}% <-this % stops a space
\thanks{Zehao Lin, Guodun~Li and  Yin~Zhang are with the College of Computer Science and Technology, Zhejiang University, Hangzhou,
Zhejiang, 310027, China. (e-mail: georgelin@zju.edu.cn, guodun.li@zju.edu.cn, and zhangyin98@zju.edu.cn).}% <-this % stops a space
\thanks{Shaobo Cui, Xiaoming Kang, Feng Ji, Fenglin Li, Zhongzhou Zhao, and Haiqing Chen are with DAMO Academy, Alibaba Group. (e-mail: yuanchun.csb@alibaba-inc.com, kxm180043@alibaba-inc.com, zhongxiu.jf@alibaba-inc.com, fenglin.lfl@alibaba-inc.com, zhongzhou.zhaozz@alibaba-inc.com and haiqing.chenhq@alibaba-inc.com).}% <-this % stops a space
\thanks{Manuscript received March 12, 2021; revised July 14, 2021.}}

\markboth{Journal of \LaTeX\ Class Files,~Vol.~14, No.~8, August~2015}%
{Shell \MakeLowercase{\textit{et al.}}: Bare Demo of IEEEtran.cls for IEEE Journals}

\maketitle

\begin{abstract}
Different people have different habits of describing their intents in conversations. Some people tend to deliberate their intents in several successive utterances, i.e., they use several consistent messages for readability instead of a long sentence to express their question. 
This creates a predicament faced by the application of dialogue systems, especially in real-world industry scenarios, in which the dialogue system is unsure whether it should answer the query of user immediately or wait for further supplementary input. 
Motivated by such an interesting predicament, we define a novel \textbf{Wait-or-Answer} task for dialogue systems. We shed light on a new research topic about how the dialogue system can be more intelligent to behave in this Wait-or-Answer quandary.
Further, we propose a predictive approach named \textbf{Predict-then-Decide~(PTD)} to tackle this \textit{Wait-or-Answer} task. 
More specifically, we take advantage of a \textit{decision} model to help the dialogue system decide whether to wait or answer. The decision of decision model is made with the assistance of two ancillary prediction models: a user prediction and an agent prediction. 
The user prediction model tries to predict what the user would supplement and uses its prediction to persuade the decision model that the user has some information to add, so the dialogue system should wait. 
The agent prediction model tries to predict the answer of the dialogue system and convince the decision model that it is a superior choice to answer the query of user immediately since the input of user has come to an end. We conduct our experiments on two real-life scenarios and three public datasets.
Experimental results on five datasets show our proposed PTD approach significantly outperforms the existing models in solving this Wait-or-Answer problem. 
\end{abstract}

\section{Introduction}
With the availability of large-scale dialogue corpora and the advances in deep learning and reinforcement learning, conversational artificial intelligence has made great progress. In recent years, many works try to improve the performance of data-driven dialogue systems from multiple dimensions, e.g. dialogue states \cite{Wu2019TransferableMS,Le2020NonAutoregressiveDS}, dialogue generation \cite{Li2017AdversarialLF}, emotion integration \cite{Hasegawa2013PredictingAE}, knowledge integration \cite{ghazvininejad2018knowledge,wu2019global}.

In real-life scenarios, we observe that a large portion of dialogue system users often describe their intents in several successive utterances rather than a single utterance. This brings a critical dilemma in which the dialogue system is not sure whether it should \textbf{wait} for the further input from the user or simply \textbf{answer} the question right away. 
If dialogue systems can not solve wait-or-answer dilemma, users have to convey their intents in a single turn, which harms the user experience. Additionally, too early cut-in or delayed response of dialogue systems will puzzle the users and lead to conversation failure. 
This Wait-or-Answer dilemma becomes even more complicated and complex when it comes to multi-turn dialogue systems. Despite the surge of attention into the dialogue system models, very few research works have specially investigated the Wait-or-Answer problem. Some studies have been done on the incomplete user utterance task \cite{DBLP:journals/corr/abs-2008-01474,DBLP:conf/emnlp/PanBWZL19,DBLP:conf/emnlp/LiuCLZZ20}, however, in multi-turn dialogue environment, all utterances from users are complete sentences (i.e. all sentences are meaningful and unbroken in both syntax and semantics). This requires our model to be able to learn user intents and habits from context. 

\begin{figure}
\centering
  \includegraphics[width=1\linewidth]{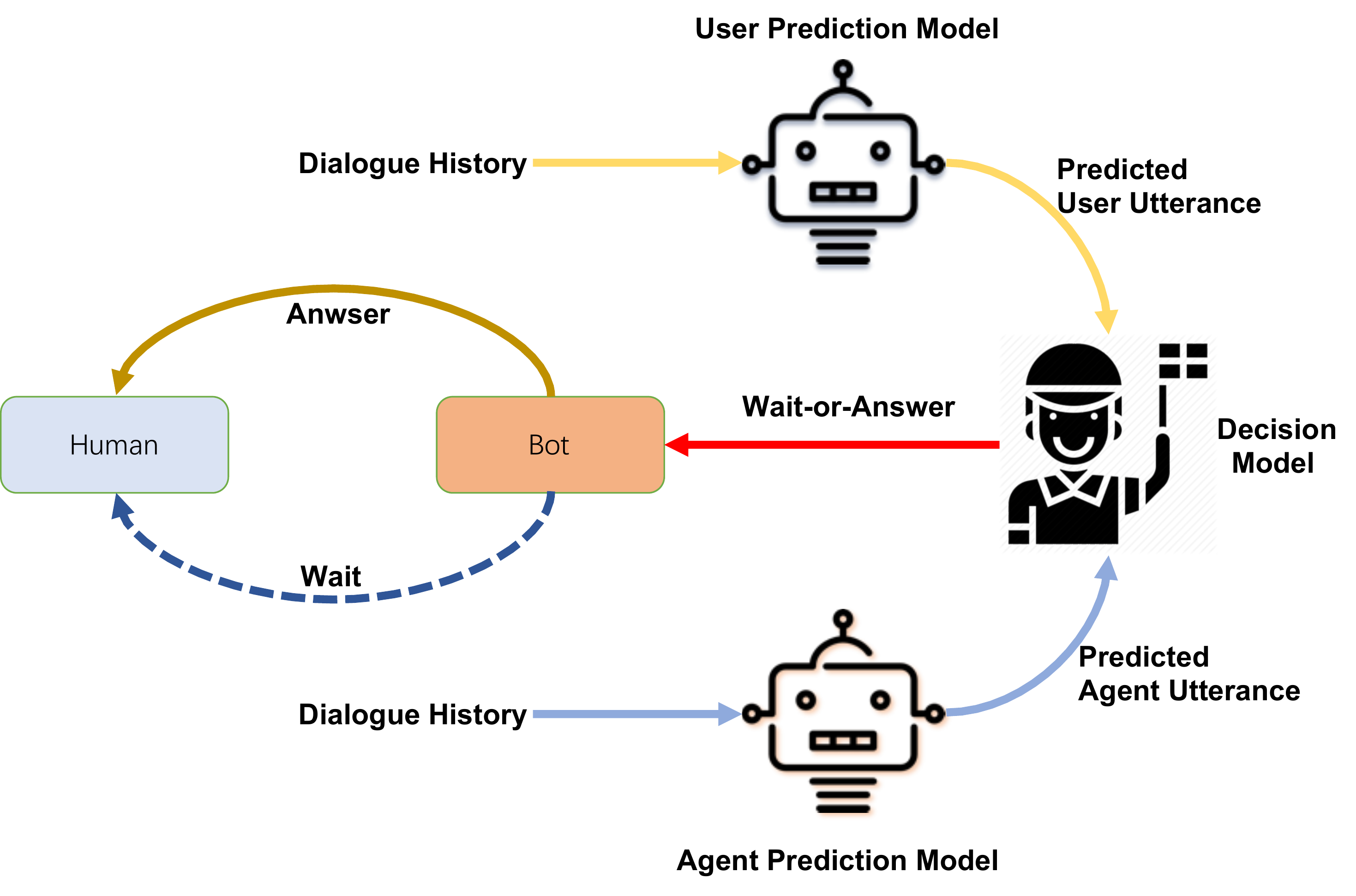}
  \caption{An overview of the PTD framework. The user prediction model and agent prediction model predict user's and agent's future possible utterance, respectively, and assist the decision model in solving \textit{Wait-or-Answer} problem.}
    \label{fig:model_overview}
\end{figure}%

The most widely-applied approach to work around in industry \cite{oard2012query,nordlie1999user} is to extend dialogue systems' response time: the dialogue agents always wait for some time in case of user's further input. However, this \textit{passive} extended-waiting-time strategy may cause incoherence of the conversation and lead to poor user experience. For example, a 3-second extended-waiting-time may seem short to some people but apparently too long for some other users. A comfortable waiting time varies with each individual, which makes it difficult to set the accurate waiting time. Recently, Liu et al. \cite{DBLP:conf/kdd/LiuJXYY20} address the similar problem under 
the semantic matching task and obtain the
weights of the co-occurring utterances in the context to determine whether the given context should be responded to.

Different from the previous strategy, we propose to address the Wait-or-Answer issue in a
\textit{predictive approach}, which purely relies on dialogue context information and can solve wait-or-answer task alone or as a supplement to time expanding and triggering methods. 
Specifically, we try to predict the user's next action based on the current dialogue history: (1) to provide supplemental information~(if so, the bot should wait). (2) to wait for the response from chatbot~(if so, the bot should answer the user's query).
The most intuitive predictive approach is to apply a classification model, which we use is a classification model to predict whether the dialogue system should wait or answer based on the dialogue history. 
These kinds of methods only consider the information in past dialogue history but omit the user and agent's possible future intention. Intuitively, suppose that we can predict what the user would supplement if the user wants to express further information, and what the dialogue system would answer if the user has completed his or her question and is waiting for an answer, the dialogue system has more confidence to decide whether to wait or answer.

Motivated by such intuitions, we propose a model named \textbf{Predict-then-Decide}~(PTD). As shown in Figure~\ref{fig:model_overview}, there is a decision model that controls whether the bot should answer the user query or wait for further information. Except for the decision model, there are two auxiliary prediction models: the user prediction and the agent prediction. The user prediction model persuades the decision model that the bot should wait for further input from users. While for the agent prediction model, it tries to convince the decision model that the bot should immediately answer users' queries. 
As for the decision model, given the \textit{suggestions} from these two prediction models, it makes its decision whether the bot should wait or answer. 

At last, to evaluate the performance of PTD framework, we test several baselines including two rule-based systems and three supervised models, and three PTD variants on two real industry datasets collected by ourselves and three public datasets for better reliability. Experimental results and analysis show the improvements brought by PTD framework.

In summary, this paper makes the following contributions:

\begin{itemize}
    \item This paper explicitly defines the Wait-or-Answer task, which is crucial to further enhance the capability of dialogue systems;
    \item  We propose a novel framework, named Predict-then-Decide~(PTD), to solve the Wait-or-Answer task, which uses two prediction models and one decision model to help dialogue systems decide whether to wait or answer; 
    \item  Experimental results on both real industry scenarios and public datasets demonstrate that our model significantly outperforms the baselines, which validates the benefits brought by our PTD framework. Our modified public datasets and code are released \footnote{Open Source Repository: https://github.com/mumeblossom/PTD} to the public for further research in both academia and industry.
\end{itemize}% 

\section{Preliminary} \label{sec:background}
In this section we provide some background knowledge about Dialogue Systems in Section~\ref{sec:related:dialogue_systems}. 
Besides, our proposed PTD framework involves both generative models~(for prediction models in PTD) and classification models~(for the decision model in PTD), we present some preliminary work about the generative and classification models in NLP in Section~\ref{sec:related:models}. 
\subsection{Dialogue Systems} \label{sec:related:dialogue_systems}
%Creating a perfect artificial human-computer dialogue system is always the ultimate goal of natural language processing. 
Researches on dialogue systems are mainly divided into two categories: task-oriented dialogue systems and chit-chat dialogue systems. Task-oriented dialogue systems \cite{DBLP:conf/eacl/ManningE17,DBLP:conf/sigdial/EricKCM17, YanDCZZL17} aim at solving tasks in specific domains with grounding knowledge while chit-chat bots \cite{yan2018chitty, li2019follow, hancock2019learning} mainly concentrate on interacting with a human to provide reasonable responses and entertainment \cite{DBLP:journals/sigkdd/ChenLYT17}. 
Recent years research on task-oriented dialogue systems mainly concentrates on dialogue states \cite{budzianowski-etal-2018-multiwoz} and knowledge integration \cite{wu2019global, lin-etal-2019-task} using pipeline or end to end models. Chit-chat bots focus on conversing with the human in open domains. Though chit-chat bots seem to perform totally different from task-oriented dialogue systems, actually as revealed in Yan et al. \cite{yan2017building}, nearly 80\% utterances are chi-chat messages in the online shopping scenario and handling those queries is closely related to user experiences. 
%The recent development of big data and deep learning techniques has greatly advanced both task-oriented dialogue systems and chit-chat bots, which has encouraged a huge amount of deep learning based researches in dialogue systems. 
Many studies have investigated how to apply neural networks to the components of dialogue systems or end-to-end dialogue frameworks \cite{YanDCZZL17,lipton2018bbq-networks}. The advantage of deep learning is its ability to leverage large amounts of data from the internet, sensors, etc. The big conversation data and deep learning techniques like SEQ2SEQ \cite{NIPS2014_5346} and attention mechanism \cite{DBLP:conf/emnlp/LuongPM15} help the model understand the utterances, retrieve background knowledge and generate responses.

\subsection{Generative and Classification Models}
\label{sec:related:models}

\noindent \textbf{Dialogue Generation} \quad
In general, two major approaches have been developed for dialogue divided by the reply types: generative methods such as sequence-to-sequence models, which generate proper responses during the conversation; and retrieval-based methods, which learn to select responses from the current conversation from a repository. 

The generative method has been attracting more and more attention \cite{Liu2018KnowledgeDF, Zhao2020LowResourceKD}. The main reason is that, when compared to retrieval-based dialogue systems, generative models can sometimes produce more fluent and flexible replies, making them more user friendly in some cases. Unlike the retrieval method, Natural Language Generation (NLG) translate a communication goal selected by the dialogue manager into a natural language form~\cite{gao2019neural}. It reflects the naturalness of a dialogue system, and thus the user experience. Another reason is that, in addition to the fluency and accuracy of responses, generative systems are far more flexible to use for common users than retrieval based systems. 

Conventional template or rule-based approaches mainly contain a set of templates, rules, and hand-crafted heuristics designed by domain experts. This makes it labor-intensive yet rigid, motivating researchers to find more data-driven approaches \cite{ghazvininejad2018knowledge, lin-etal-2019-task} to optimize a generation module from corpora, one of which, Semantically Controlled LSTM (SC-LSTM) \cite{wen2015semantically}, a variant of LSTM \cite{hochreiter1997long}, gives semantic control on language generation with an extra component. As for the fully-data driven dialogue systems, SEQ2SEQ \cite{NIPS2014_5346} based encoder-decoder frameworks and attention mechanisms \cite{DBLP:conf/emnlp/LuongPM15} are still the most widely adopted \cite{lin-etal-2019-task, ghazvininejad2018knowledge, DBLP:conf/www/ChenRTZY18} techniques. Transformer~\cite{vaswani2017attention} based models, e.g. BART~\cite{lewis2020bart}, T5~\cite{DBLP:journals/jmlr/RaffelSRLNMZLL20} and GPT~\cite{NEURIPS2020_1457c0d6}, show its effectiveness compared with traditional CNN or RNN based models in generation tasks. They are able to utilize a large amount of natural language data from the internet by self-supervised learning and fine-tune themselves in downstream tasks.

\noindent \textbf{Text Classification} \quad
Text classification is a critical problem in all NLP tasks, and it has been widely investigated and studied \cite{jiang2018text,kowsari2017hdltex,lai2015recurrent,kowsari2019text} in recent decades. Text classification can be done on multiple levels, including document classification \cite{yang2016hierarchical,manevitz2001one}, sentence classification \cite{komninos2016dependency}, emotion classification \cite{xia-ding-2019-emotion}, and so on.

Though end-to-end methods play a more and more important role in dialogue system, the text classification modules \cite{jiang2018text,kowsari2017hdltex} remain very useful in many problems like emotion recognition \cite{song-etal-2019-generating}, gender recognition \cite{hoyle-etal-2019-unsupervised}, intent detection~\cite{yan-etal-2020-intent}, etc. There have been several widely used text classification methods proposed, e.g. Recurrent Neural Networks (RNNs) and CNNs. Typically RNN is trained to recognize patterns across time, while CNN learns to recognize patterns across space. \cite{kim2014convolutional} proposed TextCNNs trained on top of pre-trained word vectors for sentence-level classification tasks and achieved excellent results on multiple benchmarks.

Besides RNNs and CNNs, a powerful network architecture called Transformer \cite{vaswani2017attention} is solely based on attention mechanisms and achieves promising performance in many NLP tasks. To make the best use of unlabeled data, Devlin et al.~\cite{devlin2018bert} introduce a new language representation model with two auxiliary pre-training tasks called BERT, which stands for Bidirectional Encoder Representations from Transformers.

\section{The Wait-or-Answer Task} \label{sec:task}
\begin{figure}
\centering
  \includegraphics[width=1\linewidth]{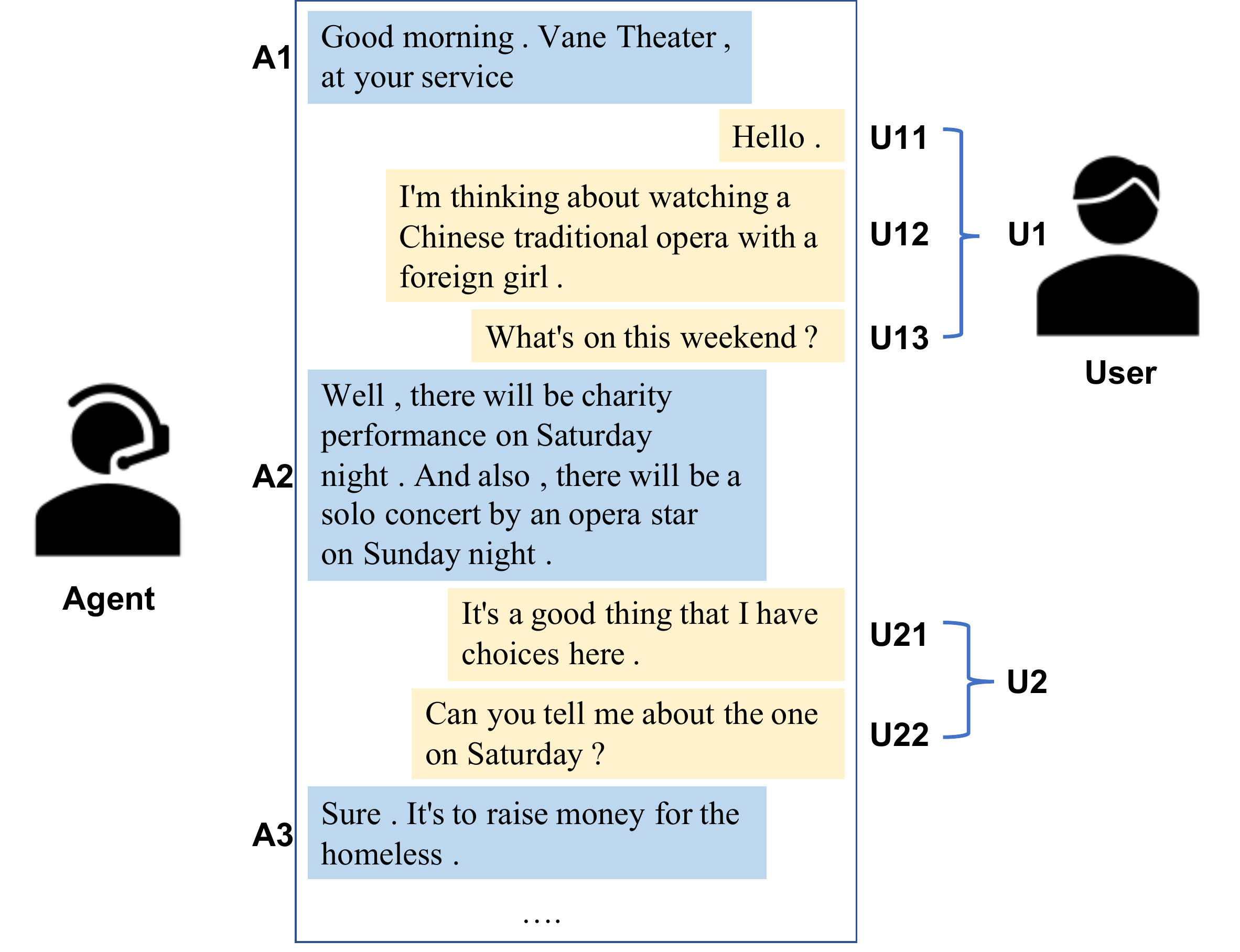}
  \caption{A multi-turn dialogue fragment. In this case, a user sends split utterances in a turn, e.g. split U1 to \{U11, U12 and U13\}}
  \label{fig:dialogue_overview}
\end{figure}

\textbf{Why Do We Study The Wait-or-Answer Task?} \quad
Conventional dialogue systems mainly concentrate on the accuracy and fluency of generated or retrieved answers. These kinds of dialogue systems, including most commercial chatbots, require users to strictly follow the designed conversation instructions.
%\replaced{
For example, users must be ready to finish all of the words they want to say in one breath and without pausing. This requires users describing their intents in a single sentence. 

However, the aforementioned setting in existing dialogue systems does NOT held in real-life settings. For instance, as shown in Figure \ref{fig:dialogue_overview},
in a real-world scenario where a user requests information about a theater, the agent firstly starts the conversation with ``Good morning. Vane Theater at your service."~(A1), then the user replies with three sentences, firstly ``Hello"~(U11), secondly ``I'm thinking about watching a Chinese traditional opera with a foreign girl."~(U12) and thirdly ``What's on this weekend?"~(U13). Generally speaking, users won't speak all sentences without a breath. If the agent cuts in the wrong moment of the conversation, e.g. immediately replies to the user's second statement, the agent has to guess what the user wants and omit the important information ``on this weekend" in the third sentence. So in this case, the agent should wait for the user until he or she has finished sending his or her last message, otherwise, the pace of the conversation will be messed up. However, existing dialogue agents can not handle this scenario well and will reply to the user's every utterance immediately or after a fixed time interval.

There are mainly two issues when applying existing dialogue agents to the real-life conversation: 
    (1) Existing dialogue systems lack the capability of deciding to avoid generating bad responses based on semantically incomplete utterance, when they receive a short utterance from users as the start of a conversation.
    (2) Existing dialogue systems may cut into a conversation at an inappropriate point, which could confuse the user and mess up the pace of conversation and thus lead to nonsense interactions. In other words, the existing dialogue system can NOT catch the right moment to Answer or Wait. 

As stated above, it is worthwhile to investigate this Wait-or-Answer task which would empower the dialogue system to enhance their ability to make appropriate decisions in the wait-or-answer dilemma.
\begin{figure*}
\centering
\includegraphics[width=1\linewidth]{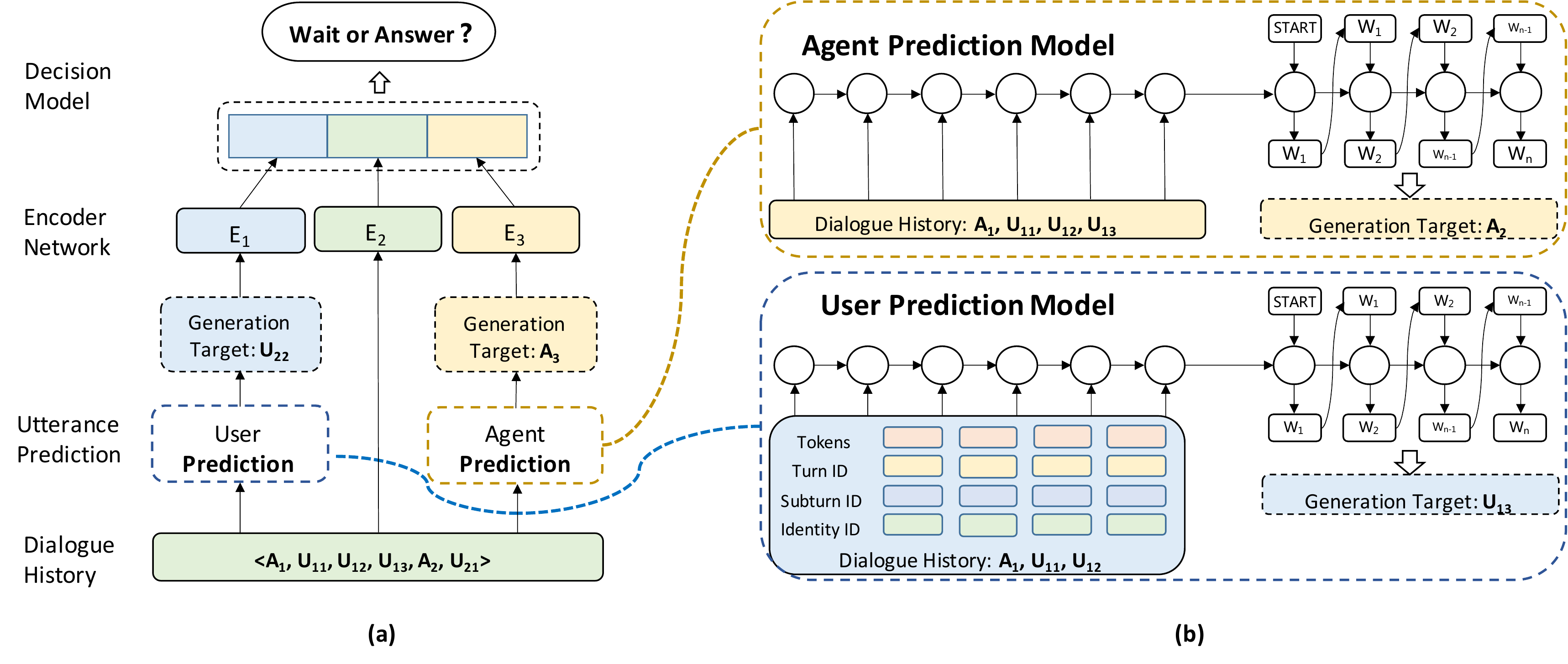}
    \caption{(a) The PTD framework will complete \textit{Wait-or-Answer} task using dialogue history and two trained prediction models' predictions.(b) Details of the agent and user prediction models trained using the same dialogues but different samples.}
\label{fig:model}
\end{figure*}   

\noindent \textbf{Task Formulation} \quad
\iffalse

\fi
The \textit{passive} extended-waiting-time strategy may cause incoherence of the conversation and lead to poor user experience. So our task is to actively predict whether the agent should answer immediately. Under this premise, our problem is formulated as follows. There is a conversation history represented
as a sequence of utterances: $X = \{x_1, x_2, ..., x_m\}$, where
each utterance $x_i$ itself is a sequence
of words $x_{i_1}, x_{i_2}, x_{i_3}...x_{i_n}$. In addition, each utterance has following additional labeled tags:

\noindent(1). {Turn id}:  which turn the utterance locates in. 

\noindent(2). {Sub-turn id}: the position of the utterance in its turn which may have more than one utterance. 

\noindent(3). {Speaker id}: who sends out the utterance. 0 means the user while 1 means the agent. 

Now, given a dialogue history $X$ and tags $T$, the goal of the model is to predict a label $Y \in \{0,1\}$, the action the agent would take, where $Y = 0$ means the agent will wait for the user for the next message, and $Y = 1$ means the agent will reply immediately. Formally we will choose the action that maximizes the following probability:
\begin{equation}
    Y =\arg \max _{y} P\left(y | X, T\right).
\label{eq:problem}
\end{equation}

\section{The Predict-then-Decide~(PTD) Framework}
\label{sec:ita}
In this section, we firstly present the overview of our PTD framework in Section~\ref{sec:ita:overview}. Section~\ref{sec:ita:imaginator} and Section~\ref{sec:ita:arbitrator} are about the detailed model structures of prediction model and decision model respectively. At last, we describe the overall training and inference phase in Section~\ref{sec:ita:phase}.

\subsection{The Overview of PTD Framework} \label{sec:ita:overview}

 The PTD framework has two prediction models and one decision model, as shown in Figure~\ref{fig:model_overview}. The decision model makes the final decision about whether the dialogue system should answer users' queries immediately or wait for the users' further information. We use two prediction models to assist the decision model. 
The user prediction model forecasts what the user might supplement. Then, the user prediction model uses this \textbf{simulated} query to convince the decision model to wait for the user's following input since the user does have some information to supplement.  
The agent prediction model, nevertheless, predicts the dialogue system's answer for users' present query. Then, the agent prediction model utilizes this \textbf{simulated} answer to make the decision model believe that the dialogue system should answer the user's queries immediately because the user has finished its input. 

In fact, these two prediction models act as the world model \cite{ha2018recurrent}, which creates a virtual environment to simulate the possible future dialogue to train the agent, for the decision model. More specifically, the output of the user prediction model~(simulated query) and the output of the agent prediction model~(simulated answer) both function as the simulated experience. Peng et al.~\cite{peng2018deep} first propose Deep Dyna-Q, incorporating into the dialogue agent a world model to mimic real user response and generate simulated experience.
Compared to models that directly apply a classifier, such as TextCNN and BERT, to dialogue systems to solve the \textit{Wait-or-Answer} problem, our proposed prediction models are better at learning semantic information, from both history and future possible utterances, by training on the corpus, and providing supplemental prediction to the decision model. Prediction models will also magnify the errors, providing negative feedback and making it easier for our PTD to understand which decision is better.

\subsection{Prediction Model}
\label{sec:ita:imaginator}
The prediction model in PTD generates the next possible utterance given the dialogue history. There are two prediction models in our method: the user prediction model and the agent prediction model. The goal of the two prediction models is to learn the user’s and agent’s speaking style respectively and generate possible future utterances from different perspectives. 

As shown in Figure \ref{fig:model} \textbf{(a)}, prediction model itself is a sequence generation model. We use one-hot embedding to convert all words and the related tags to one-hot vectors $w_n \in \textbf{R}^{|V|}$, where $|V|$ is the length of the vocabulary list. Then we extend each word $x_{i_j}$ in utterance $x_i$ by concatenating the token itself with turn tag, identity tag, and sub-turn tag. We adopt SEQ2SEQ as the basic architecture and LSTMs as the encoder and decoder networks.  
%Then we feed the conversation history $X=\{x_1, x_2, ..., x_m\}$ into a sequence of words denoted as $S=\{w_1, w_2...w_n\}$, which  
LSTMs will encode each extended word $w_t$ as a continuous vector $h_t$ at each time step $t$. The process can be formulated as: $h_{t}=\text{LSTMs}( h_{t-1} )$.
For the same piece of dialogue, we split it into different samples for different prediction models. As shown in Figure \ref{fig:dialogue_overview} and \ref{fig:model} \textbf{(a)}, we use (A1, U11, U12) as dialogue history input and U13 as ground truth to train the user prediction model and use (A1, U11, U12, U13) as dialogue history and A2 as ground truth to train the agent prediction model.

Apart from the extra labels, the prediction models are trained in the conventional way: $h_{t}=\text{LSTMs}( h_{t-1} )$, and the decoder is the similarly structured LSTMs but $h_t$ will be fed to a Softmax with $W_{v} \in {\textbf{R}^{h \times |V|}}, b_{v} \in{\textbf{R}^{|V|}}$, which will produce a probability distribution $p_{t}$ over all words, formally: $p_{t}= \text{Softmax}(W_{v} h_{t}+b_{v})$.

Decoder will select the word based on $p_{t}$ at each time step. The loss for prediction model is the sum of the negative log-likelihood of the correct word at all time steps:
\begin{equation}
\label{eq:train-img}
L_{\text{PRE}}=-\sum_{t=1}^{N}\log(p_{t}),
\end{equation}
where $N$ is the length of the generated sentence. During inference, we also apply a beam search to improve generation performance. Finally, the trained agent prediction model and user prediction model are obtained.

\subsection{Decision Model}
\label{sec:ita:arbitrator}

The decision module is fundamentally a text classifier. In our settings, to fully utilize the dialogue history and latent semantic information, we rewrite the objective in Equation~(\ref{eq:problem}) as follows:
\begin{equation}
\begin{array}{l}
\begin{aligned}
    R' =\arg \max _{y} P\left(y | X, T, R_{a}, R_{u} \right),
\end{aligned}
\end{array}\label{eq:new_problem}
\end{equation}
where ${R}_{a} = \text{IG}_{a}(X, T)$ and ${R}_{u}  = \text{IG}_{u}(X, T)$ represent the generated utterances from the agent prediction model and user prediction model respectively. $R' \in {[0, 1]}$ is a selection indicator where $R' = 1$ means selecting $R_{a}$ whereas $0$ means selecting $R_{u}$. 
%And Thus we (1) introduce the supervise information in imaginators' training data and future possible predicted utterances (2) turn the label prediction problem into a response selection problem.

We adopt several architectures like Bi-GRUs, TextCNNs, and BERT as the base model of the decision module. Without loss of generality, here we illustrate how to build a decision model by taking TextCNNs as an example.

As shown in Figure \ref{fig:model}, the three similarly structured TextCNNs following the work \cite{kim2014convolutional} take the inferred responses $R_{a}$, $R_{u}$ and dialogue history $X$, tags $T$. For each raw word sequence $x_1,...,x_n$, we encode each word as one-hot vector $w_{i} \in \textbf{R}^{|V|}$. By looking up a word embedding matrix $E \in \textbf{R}^{|V| \times d}$, the input text is represented as an input matrix $Q \in \textbf{R}^{l \times d}$, where $l$ is the length of sequence of words and $d$ is the dimension of word embedding features. The matrix is then fed into the similar structured CNNs using one layer convolution with max-over-time pooling to get the feature maps of $X$, $R_{a}$ and $R_{u}$:

\begin{equation}
\begin{aligned}
\mathrm{\hat{C}}_{x}&=\text{TextCNNs}(X); \\
\mathrm{\hat{C}}_{a}&=\text{TextCNNs}(R_{a}); \\
\mathrm{\hat{C}}_{u}&=\text{TextCNNs}(R_{u}). \\
\end{aligned}
\label{eq:textcnns}
\end{equation}
Then we will have two possible dialogue paths, $X$ with $R_{a}$ and $X$ with $R_{u}$, representing $\mathrm{D}_{a}$ and $\mathrm{D}_{u}$:
\begin{equation}
    \begin{cases}
    \mathrm{D}_{a} = W_{1}[\mathrm{\hat{C}}_{x}; \mathrm{\hat{C}}_{a}] + b_{1} & \text{\textit{Answer} path: $X$ with $R_{a}$}.\\
\mathrm{D}_{u} = W_{2}[\mathrm{\hat{C}}_{x}; \mathrm{\hat{C}}_{u}] + b_{2}  & \text{\textit{Wait} path:  $X$ with $R_{u}$}.
    \end{cases}
\end{equation}

Then the decision model will predict which of these two paths~(user path or agent path) should be taken. Namely, the decision model conducts a binary classification task: 
\begin{equation}
\begin{aligned}
P = \text{Softmax}(W_{4}(W_{3}[\mathrm{D}_{a}; \mathrm{D}_{u}] + b_{3})+b_{4}),
\end{aligned}
\end{equation}
where $W_{1}$ to $W_{4}$ and $b_{1}$ to $b_{4}$ are learnt parameters. At last we will get a two-dimensional probability distribution $P$, which indicates the most reasonable response.
The loss function of the decision model is the negative log-likelihood of the probability of choosing the correct action:
\begin{equation}
\label{eq:train-arb}
    L_{\text{DEC}} = -\sum_{i=1}^{M}\sum_{j=0}^{1}Y_i(j){\log(P(j))},
\end{equation}
where $j = \{0,1\}$ is the action label, $M$ is the number of samples and $Y_{i}$ is an one-hot encoding of the ground-truth label of the i-th sample.

The decision modules based on Bi-GRU or BERT are implemented similarly to TextCNNs in
Equation~(\ref{eq:textcnns}).

\begin{algorithm}
\caption{Training and inference of PTD Framework}
\label{alg:ita-training-process}
\begin{algorithmic}[1]
\Procedure{TRAIN-USER-PREDICTION($X$, $T$)}{}
    \State \textbf{for} mini-batch($X$, $T$) \textbf{do}
    \State \quad $R_u\leftarrow IG_u(X,T)$
    \State \quad minimize Equation~(\ref{eq:train-img}) to optimize user prediction model
	\State \textbf{end}
\State \textbf{return} trained user prediction model $IG_u$
\EndProcedure
\Procedure{TRAIN-AGENT-PREDICTION($X$, $T$)}{}
    \State \textbf{for} mini-batch($X$, $T$) \textbf{do}
    \State \quad $R_a\leftarrow IG_a(X,T)$;
    \State \quad minimize Equation~(\ref{eq:train-img}) to optimize agent prediction model
	\State \textbf{end}
\State \textbf{return} trained agent prediction model $IG_a$
\EndProcedure
\Procedure{TRAIN-DECISION($X$, $T$)}{}
    \State \textbf{for} mini-batch($X$, $T$) \textbf{do}
    \State \quad $R_u \leftarrow IG_u(X, T)$
    \State \quad $R_a \leftarrow IG_a(X, T)$
    \State \quad $R' \leftarrow$ decision model$(X, T, R_u, R_a)$
    \State \quad minimize Equation~(\ref{eq:train-arb}) to optimize decision model
    \State \textbf{end}
\State \textbf{return} trained decision model
\EndProcedure
\Procedure{INFERENCE($X$, $T$)}{}
    \State \quad $R_u \leftarrow IG_u(X, T)$
    \State \quad $R_a \leftarrow IG_a(X, T)$
    \State \quad $R' \leftarrow$ decision model$(X, T, R_u, R_a)$
\EndProcedure
\State TRAIN-USER-PREDICTION;
\State TRAIN-AGENT-PREDICTION;
\State TRAIN-DECISION;
\State INFERENCE;
\end{algorithmic}
\label{al:train}
\end{algorithm}

\subsection{Training and Inference of PTD Framework} \label{sec:ita:phase}

\begin{table*}[]
\centering
\caption{Real Industry Datasets Statistics.}
\label{tab:real_data_stastics}
\resizebox{0.7\linewidth}{!}{%
\begin{tabular}{lcccccc}
\hline
Datasets                             & \multicolumn{3}{c}{E-COMM} & \multicolumn{3}{c}{After-Sales}\\ \hline
Split                                & Train    & Valid   & Test    & Train     & Valid     & Test\\ \hline
Vocabulary Size                      & \multicolumn{3}{c}{9948}     & \multicolumn{3}{c}{3877} \\
Dialogues                            & 1676     & 209    & 211    & 1796     & 224      & 226      \\
Avg. Turns/Dialogue                  & 12.50     &  11.79   &  12.03   &   10.14    &   8.76    & 11.26      \\
Avg. User Sub-turns                & 1.40     & 1.41    & 1.40    & 1.56      & 1.59      & 1.55      \\
Avg. Utterance Length                & 41.37    & 45.27    & 40.75   & 50.07      & 49.78      & 47.04     \\
Avg. Agent's Utterances Length & 28.78    & 29.39   & 28.83   & 24.49     & 24.40     & 24.13     \\
Avg. User's Utterances Length  & 5.69     & 8.08    & 6.10    & 8.84      & 8.54      & 7.72     \\
Agent Wait Samples Size              & 20976    & 2464    & 2538    & 18213     & 1962      & 2544     \\
Agent Answer Samples Size            & 8348    & 1016    & 1027    & 10112     & 1150      & 1401      \\ \hline
\end{tabular}%
}
\end{table*}

\begin{table*}[]
\centering
\caption{Public Datasets Statistics. Note that the statistics are based on the modified dataset described in Section \ref{p:d_m}}
\label{tab:data_stastics}
\resizebox{1\linewidth}{!}{%
\begin{tabular}{lccccccccc}
\hline
Datasets                             & \multicolumn{3}{c}{MultiWOZ} & \multicolumn{3}{c}{DailyDialog} & \multicolumn{3}{c}{CCPE} \\ \hline
Split                                & Train    & Valid   & Test    & Train     & Valid     & Test      & Train   & Valid  & Test  \\ \hline
Vocabulary Size                      & \multicolumn{3}{c}{2443}     & \multicolumn{3}{c}{6219}          & \multicolumn{3}{c}{4855} \\
Dialogues                            & 8423     & 1000    & 1000    & 11118     & 1000      & 1000      & 398     & 49     & 52    \\
Avg. Turns/Dialogue                  & 6.32     & 6.97    & 6.98    & 4.09      & 4.21      & 4.03      & 9.7     & 9.96   & 9.92  \\
Avg. Split User Turns                & 1.89     & 1.92    & 1.94    & 2.09      & 2.12      & 2.12      & 3.12    & 3.02   & 2.73  \\
Avg. Utterance Length                & 10.54    & 10.7    & 10.56   & 8.71      & 8.54      & 8.75      & 7.93    & 8.02   & 7.78  \\
Avg. Agent's Utterances Length & 14.43    & 14.78   & 14.69   & 12.04     & 11.81     & 12.17     & 8.7     & 8.84   & 8.19  \\
Avg. User's Utterances Length  & 6.18     & 6.28    & 6.17    & 5.91      & 5.87      & 5.96      & 7.61    & 7.66   & 7.56  \\
Agent Wait Samples Size              & 47341    & 6410    & 6573    & 49540     & 4717      & 4510      & 8183    & 973    & 894   \\
Agent Answer Samples Size            & 53249    & 6970    & 6983    & 41547     & 3846      & 3689      & 3455    & 436    & 464   \\ \hline
\end{tabular}%
}
\end{table*}

\begin{figure*}
    \centering
    \includegraphics[width=0.9\linewidth]{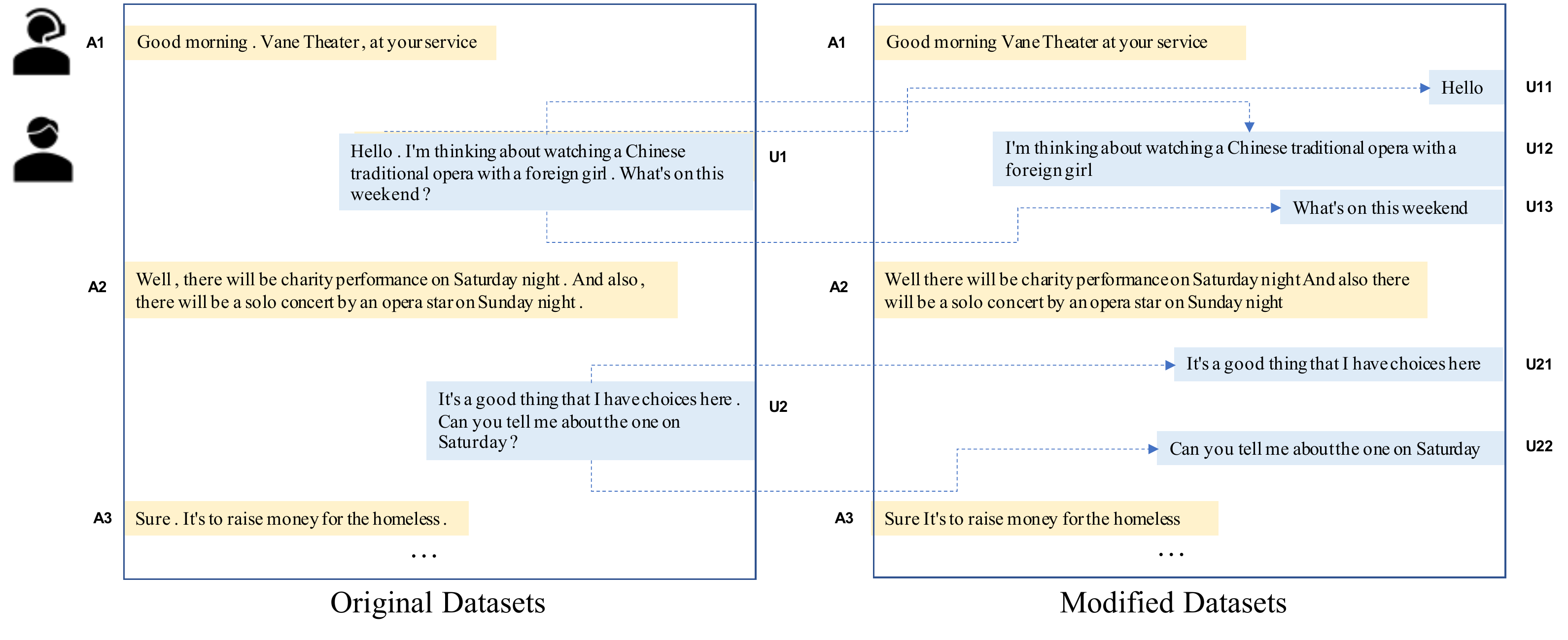}
\caption{Comparison of a piece of the original dialogue and Our modified dialogue following the section \ref{sec:dataset:construct_pipeline}. Note that some elements task-oriented corpus like MultiWOZ slot values, knowledge base, and ontology content are not shown in Original Datasets and have been deleted in Modified Datasets.}
\label{fig:datacon}
\end{figure*}

% \end{table*}

As shown in Figure \ref{fig:model} and Algorithm \ref{al:train}, we show the procedure of the model's training and inference. We first train user prediction model on the dialogue history with ground-truth as user's utterance (For example, from ~[$A_1$, $U_{11}$] to $U_{12}$ in Figure \ref{fig:dialogue_overview}) and agent prediction model on the dialogue history with ground-truth as agent's utterance (For example, from ~[$A_1$, $U_{11}$, $U_{12}$, $U_{13}$] to $A_2$). And then we infer predicted user and agent future possible utterances from user and agent prediction models as decision model's training data. In this kind of design, the two prediction models will not only simulate future possible dialogue to support \textit{wait} and \textit{answer} action, but also magnify the distinction between decisions, e.g. the performance of user prediction model will be poor when the ground-truth is \textit{answer} because user prediction model never learned how to speak like an agent, this will make decision model easier to distinguish which decision is better. At last, we feed the predicted utterances and original dialogue history together to train the decision model.
During the inference procedure, we simply use two prediction models to predict possible user and agent's utterance ($U_{22}'$ and $A_3'$) and combined with dialogue histories ~[$A_1$, $U_{11}$, $U_{12}$ $U_{13}$ $A_2$ $U_{21}$] for the decision model to decide whether the model should wait or answer right away.

\section{Experimental Setup} \label{sec:setup}
In this section, we firstly present the process of data construction in Section~\ref{sec:experiment:dataset}, after which we give the evaluation metrics we use in Section~\ref{sec:experiment:metrics}. Finally, we detail the training setup of baselines and our PTD models in Section~\ref{sec:experiment:baselines} and Section~\ref{sec:experiment:ita} respectively.

\subsection{Datasets Construction} \label{sec:experiment:dataset}
\subsubsection{Realistic Scenarios}
We test our approach and baselines on two real scenarios. Table \ref{tab:real_data_stastics} shows the statistics of two real industry datasets.
\begin{itemize}
    \item E-COMM. E-COMM is a dataset that contains real dialogue history between customers and service representatives from one of the biggest cross-border e-commerce. In a session of dialogue, both customers and customer service may have sub-turns. We merge the sub-turns of customers service in each turns into one long utterance.
    \item After-Sales. After-Sales is a dataset that contains real dialogue history between customers and customer service, talking about equipment maintenance and repairing topics. We take the same pre-processing method as E-COMM.
\end{itemize}
\subsubsection{Public Datasets} \label{sec:dataset:experiment:original}
As the proposed approach mainly concentrates on the interaction between human and computer, we select and modify three datasets of very different styles to evaluate the performance and generalization of our method. Two of them are task-oriented dialogue datasets. One is a large  MultiWOZ 2.0\footnote{http://dialogue.mi.eng.cam.ac.uk/index.php/corpus/} and the other is a smaller dataset Coached Conversational Preference Elicitation (CCPE)\footnote{https://research.google/tools/datasets/coached-conversational-preference-elicitation/}, which has many more turns per dialogue. 
 The last dataset is a chit-chat dataset DailyDialog\footnote{http://yanran.li/dailydialog.html}. 
  All datasets are collected during human-to-human conversations. We evaluate and compare the results with the baseline methods from multiple perspectives. Table \ref{tab:data_stastics} shows the statistics of datasets and details of datasets are described as follows:

\begin{itemize}[leftmargin=*]
    \item \emph{MultiWOZ 2.0} \cite{budzianowski-etal-2018-multiwoz}. MultiDomain Wizard-of-Oz dataset (MultiWOZ) is a
    fully-labeled collection of human-human written conversations. Compared with previous task-oriented dialogue datasets, e.g. DSTC 2 \cite{henderson-etal-2014-second} and KVR \cite{DBLP:conf/sigdial/EricKCM17}, it is a much larger multi-turn conversational corpus and across several domains and topics. 
    \item \emph{DailyDialog} \cite{li-etal-2017-dailydialog}. DailyDialog is a high-quality multi-turn dialogue dataset, which contains conversations about daily life. In this dataset, humans often first respond to the previous context and then propose their own questions and suggestions. In this way, people pay more attention to others’ words and are willing to continue the conversation. The speaker's behavior will be more unpredictable and complex than in the task-oriented dialogue datasets.
    \item \emph{CCPE} \cite{radlinski2019coached}. CCPE is a dataset consisting of 502 English dialogues. Though it seems much smaller than MultiWOZ 2.0 and DialyDialogue, CCPE has 12,000 annotated utterances between a user and an assistant discussing movie preferences in natural language. It is collected using a Wizard-of-Oz methodology between two paid crowd-workers and focuses on the movie domain. We select this dataset to test if our model can run well on both large and small datasets.
\end{itemize}

\subsubsection{The Pipeline of Dataset Construction}  \label{sec:dataset:construct_pipeline}
\label{p:d_m}

\begin{table*}[]
\centering
\caption{An Example of The Prediction Model's Generation and decision model's Selection. Decision model chooses the agent prediction model means agent should answer the user directly.}
\label{tab:example}
\resizebox{0.7\linewidth}{!}{%
\begin{tabular}{lll}
\hline
\multicolumn{3}{c}{Example}                                                                         \\ \hline
\multirow{1}{*}{Dialogue History} & \multicolumn{2}{l}{\textbf{User}: what is the address for pizza hut in cherry hinton}                                          \\ \hline
Ground-Truth                      & \multicolumn{2}{l}{\textbf{Agent: the address is [restaurant\_address]}}  \\ \hline
\multirow{2}{*}{\begin{tabular}[c]{@{}l@{}} Prediction\end{tabular}} &
  Agent Prediction Model &
  the address is [restaurant\_address]  can i help you with anything else \\
                                  & User Prediction Model            &  i am looking for a guesthouse           \\ \hline
Decision Model Selection              & \multicolumn{2}{l}{\textbf{Agent Prediction Model}}                    \\ \hline
\end{tabular}%
}
\end{table*}

\begin{table}[]
    \centering
    \caption{Accuracy Results on Two real scenario datasets. Better results between baselines and corresponding PTD models are in \textbf{bold}.}
    \resizebox{0.46\textwidth}{!}{
    \begin{tabular}{lcccc}
\toprule
Models & \multicolumn{2}{c}{E-COMM}       & \multicolumn{2}{c}{After-Sales} \\ \midrule
         & Accuracy         & F1     & Accuracy         & F1    \\ \hline
ATLU      & 39.16          & 35.74 & 43.46 & 32.91\\
PTSU      & 48.18          & 57.47& 49.80& 58.58 \\ \hline
Bi-GRU      & 71.96          & 78.99      & 64.32          & 78.23  \\
\textbf{GRU-PTD} &
  \textbf{72.06} &
  \textbf{83.69} &
  \textbf{70.85} &
  \textbf{82.62}\\ \hline
TextCNN      & 69.91          & 76.60& 65.78 & 78.63 \\
\textbf{TextCNN-PTD} &
  \textbf{72.08} &
  \textbf{83.71} &
  \textbf{70.71} &
  \textbf{82.53} \\ \hline
BERT         & 69.45 & 81.46     & 66.02          & 79.43    \\
\textbf{BERT-PTD} &
  \textbf{71.98} &
  \textbf{83.71} &
  \textbf{71.14} &
  \textbf{82.86}  \\ \bottomrule
\end{tabular}
    }

    \label{tab:rea_results}
\end{table}

As the task we concentrate on, making a decision to wait or answer, is quite different from traditional dialogue systems, existing dialogue datasets will be unable to provide the information for training and testing. Thus we propose a fairly simple and general dataset construction method to directly rebuild over the existing public dialogue corpus.

We modify the datasets with the following steps:

\begin{enumerate}[leftmargin=3mm,label=\textbf{\Roman*.}]
    \item \textbf{Delexicalisation:} For task-oriented dialogue, slot labels are important for navigating the system to complete a specific task. However, those labels and accurate values from ontology files will not benefit our task essentially. So we replace all specific values with a slot placeholder in the pre-processing step.
    \item \textbf{Utterance segmentation:} Existing datasets concentrate on the dialogue content, combining multiple sentences into one utterance each turn when gathering the data. In this step, we split the combined utterance into multiple \textit{complete} utterances according to original datasets. This makes task in line with user habits. And models can not make decisions simply based on the completeness of sentences. In this pre-processing procedure, we do not divide all user utterances but half part of all in train/development/test sets. All punctuation marks are eliminated after this procedure to improve the applicability in most situations including Spoken Dialogue Systems. 
    \item \textbf{Extra Labeling:} We add several labels, including turn tags, sub-turn tags, and role tags, to each split and original sentences in order to (1) label the speaker role and dialogue turns (2) mark the ground truth for supervised training and evaluate the baselines and our model.
\end{enumerate}

Finally, we have the modified datasets which imitate the real-life human chatting behaviors. As shown in Figure \ref{fig:datacon}, we compare one original dialogue (in this example, from DailyDialog) with our modified one. Our modified datasets and code are open-sourced to both academic and industrial communities.

\subsection{Evaluation Metrics} \label{sec:experiment:metrics}
In our \textit{Wait-or-Answer} task, we define the \textit{Answer} action of the agent as the positive samples and the \textit{Wait} action is the negative action. As both the positive and negative actions are important in this task, so we choose the model with the accuracy metrics instead of precision or recall.

To compare with dataset baselines in multiple dimensions and test the model's performance, we use the overall Bilingual Evaluation Understudy (BLEU) \cite{DBLP:conf/acl/PapineniRWZ02}, which is the cumulative score for BLEU-1 to BLEU-4, to evaluate the prediction models' generation performance. As for the decision model, we use the accuracy score as the main metrics to evaluate the wait-or-answer decision and select models. Though our aim is to obtain a model with best accuracy in distinguishing wait or answer, apart from BLEU and accuracy, we also present Precision, Recall and F1 to evaluate baselines and our models from multiple perspectives. Details are as follows:
\begin{itemize} [leftmargin=*]
\item \emph{Bilingual Evaluation Understudy (BLEU)} \cite{DBLP:conf/acl/PapineniRWZ02}. BLEU has been widely employed in evaluating sequence generation including machine translation, text summarization, and dialogue systems. BLEU calculates the n-gram precision which is the fraction of n-grams in the candidate text which is present in any of the reference texts.
% \subsubsection{arbitrator Metrics}
% \begin{itemize} 
\item \emph{Accuracy} The accuracy metric is the probability of whether the decision model can successfully classify the ground truth in the test dataset. The accuracy score in our experiments is the correct ratio in all samples. 
\item \emph{Precision} also called positive predictive value is the fraction of relevant instances among the retrieved instances. In our case, we calculate precision by the ratio of correctly predicted answer actions in all predicted answer actions of the test dataset.
\item \emph{Recall} also known as sensitivity is the fraction of the total amount of relevant instances that are actually retrieved. In our case, we calculate recall by the ratio of correctly predicted answer actions in all answer actions of the test dataset.
\item \emph{F1 Score} Only consider the precision p or the recall r is difficult to determine which one is really better of not both p and r get a better score. F1 score considers both the precision p and the recall r of the test to compute the score. We calculate the F1 score by the harmonic mean of the precision and the recall.
\end{itemize}

\begin{table*}[]
\centering
\caption{Results on Three Public Datasets. Better results between baselines and corresponding PTD models are in \textbf{bold}.}
\label{tab:f1}
\resizebox{1\linewidth}{!}{%
\begin{tabular}{lcccccccccccc}
\hline
Dataset & \multicolumn{4}{c}{MultiWOZ}       & \multicolumn{4}{c}{DailyDialog}  & \multicolumn{4}{c}{CCPE}   \\ \hline
        & Accuracy & Precision & Recall         & F1    & Accuracy & Precision & Recall         & F1    & Accuracy & Precision & Recall & F1    \\ \hline
  ATLU      & 52.57 & 60.89 & 22.18         & 32.52    & 57.71 & 54.67 & 35.21         & 42.84    & 61.27 & 35.24 & 15.95 & 21.96    \\PTSU      & 60.78 & 60.93 & 66.53         & 63.61    & 59.26 & 53.62 & 70.07         & 60.75    & 42.27 & 33.44 & 69.61 & 45.17   \\ \hline
Bi-GRU  & 79.12 & 75.69     & 87.70          & 80.85 & 75.23 & 72.07     & 72.94          & 71.86 & 67.53 & 54.83     & 31.15  & 38.49 \\
\textbf{GRU-PTD} &
  \textbf{82.03} &
  \textbf{79.02} &
  \textbf{88.87} &
  \textbf{83.27} &
  \textbf{77.80} &
  \textbf{77.97} &
  \textbf{77.29} &
  \textbf{75.71} &
  \textbf{72.69} &
  \textbf{63.40} &
  \textbf{48.47} &
  \textbf{53.50} \\ \hline
TextCNN & 77.68 & 73.61     & 88.85          & 80.03 & 75.79 & 71.03     & \textbf{78.49} & 73.91 & 68.65 & 59.78     & 27.63  & 36.35 \\
\textbf{TextCNN-PTD} &
  \textbf{80.75} &
  \textbf{77.17} &
  \textbf{89.08} &
  \textbf{82.52} &
  \textbf{79.02} &
  \textbf{77.14} &
  74.87 &
  \textbf{75.35} &
  \textbf{73.32} &
  \textbf{68.99} &
  \textbf{41.90} &
  \textbf{51.43} \\ \hline
BERT    & 80.75 & 76.93     & \textbf{89.46} & 82.73 & 78.68 &  75.03     & 78.86          & 76.90 & 70.99 & 59.21     & 48.49  & 53.31 \\
\textbf{BERT-PTD} &
  \textbf{82.73} &
  \textbf{80.36} &
  87.99 &
  \textbf{84.00} &
  \textbf{79.35} &
  \textbf{75.92} &
  \textbf{79.23} &
  \textbf{77.54} &
  \textbf{75.41} &
  \textbf{67.86} &
  \textbf{53.23} &
  \textbf{59.66} \\ \hline
\end{tabular}%
}
\end{table*}

\subsection{Baselines and Their Training Setup} \label{sec:experiment:baselines}
To make the best practice hyper-parameter settings adopted by each training set in baselines and our models. For baselines, we only rely on dialogue history to make decisions.
We conduct experiments on the following baselines with fine-tuned parameters: 
\begin{itemize}[leftmargin=*]
    \item \textbf{Gated Recurrent Units(GRU)} \cite{chung2014empirical}: we use GRU to encode the input history as vector and convert it into a two-dimensional probability vector with softmax to predict whether the model should wait or answer. We test hidden size ranging from 200 to 600, dropout rate ranging from 0.2 to 0.8, batch size in [32, 64, 128, 256]. 
    \item \textbf{TextCNN} \cite{kim2014convolutional}: we use TextCNN to encode the input history as vector and convert it into a two-dimensional probability vector with softmax to predict whether the model should wait or answer and search the best performance in batch size ranging in [32, 64, 128, 256], dropout rate from 0.3 to 0.7,  kernel number, which is the number of convolution kernels of each size type, from 100 to 600, kernel size in [(1,2,3),(3,4,5),(5,6,7),(7,8,9)]. 
    \item \textbf{BERT} \cite{devlin2018bert}: we use BERT to encode the input history as vector and convert it into a two-dimensional probability vector with softmax to predict whether the model should wait or answer and test learning rate in [2e-5, 3e-5, 5e-5], training epochs in [2.0, 3.0, 4.0] and batch size in [16, 32].
    \end{itemize}
    We aslo employ two rule-based baselines referred to the work in Liu et al.~\cite{DBLP:conf/kdd/LiuJXYY20}:
    \begin{itemize}
    \item \textbf{Active Triggering based on Longest Utterance (ATLU)}: It considers the lengths of utterances in the input history to make decisions. If the length of the last utterance is longer than others, the model should answer directly.
    \item \textbf{Passive Triggering based on Shortest Utterance (PTSU)}: Similar to ATLU, it compares the length of the last utterance with the lengths of all utterances in the input history to make decisions. The model should wait only if the length of the last utterance is shorter than others.
\end{itemize}
\subsection{PTD Models and Their Training Setup} \label{sec:experiment:ita}
To test the performance of our proposed PTD framework, we apply our PTD framework in the baselines and obtain \textbf{GRU-PTD}, \textbf{TextCNN-PTD} and \textbf{BERT-PTD}. The detailed setting on public datasets is described as follows: 
\begin{itemize}[leftmargin=*]
    \item \textbf{GRU-PTD}: for GRU-PTD on MultiWOZ, batch size is 32, hidden size is 300, dropout rate is 0.3. On DailyDialog, batch size is 64, hidden size is 500, and dropout rate is 0.5. On CCPE, batch size is 32, hidden size is 200, and dropout rate is 0.8. 
    \item \textbf{TextCNN-PTD}: for TextCNN-PTD on MultiWOZ, batch size is 64, kernel number is 400, kernel size is (7,8,9), and dropout rate is 0.3. On DailyDialog, batch size is 32, kernel number is 400, kernel size is (5,6,7), and dropout rate is 0.5. On CCPE, batch size is 64, kernel number is 600, kernel size is (5,6,7), and dropout rate is 0.4. 
    \item \textbf{BERT-PTD}: the maximum sequence length to 128, batch size is 32, and the number of training epochs is 3.0 to 4.0.
\end{itemize}

During training, we also adopt a learning rate decay factor as 0.5. All experiments employ the teacher-forcing scheme~\cite{bekey1992neural}, feeding the gold target of last time. We also perform early stopping for decision model when its performance does not increase during 6 consecutive validation epochs. We test the hidden size in [32, 64, 128, 256] and set dropout rate in [0.1, 0.2]. The learning rate is initiated with 0.001 and the training batch is set to 64. The metrics results are coming from the best result settings for each dataset. 

\section{Experimental Results and Analysis}
\subsection{Experiment Results}
\label{sec:results}

\noindent\textbf{A Case Showing How PTD Works} \quad
{For a better understanding of our PTD framework, we also present an example of how PTD acts in Table \ref{tab:example}. The User prediction model predicts what the user might supplement based on the dialogue history: \textit{i am looking for a guesthouse}. 
The agent prediction model, however, predicts what the dialogue system might answer: \textit{
  the address is [restaurant\_address]  can i help you with anything else}. Based on both predictions, the decision model concludes that it's a better choice to take the supplementary input of agent. So the dialogue system decides to continue the dialogue rather than to wait. }

\noindent\textbf{Experimental Results} \quad
To illustrate the benefits brought by our PTD framework, we present the comparison results between our PTD models~\footnote{Without loss of generality, the prediction models in our PTD models: GRU-PTD, TextCNN-PTD, BERT-PTD all adopt an LSTM structure applied with the attention mechanism.} and the baselines on real industry datasets in Table~\ref{tab:rea_results}. We can observe that our PTD models achieve superior performance compared with their counterparts. Additionally, to evaluate our models and baselines from other multiple dimensions, we present the comparison results~\footnote{These models are selected by the accuracy scores} on accuracy, precision, recall, and F1 results in Table \ref{tab:f1}, which also prove the benefits brought by our PTD framework. To better analyze the effects of prediction models in our PTD framework, we present different PTD models' performance scores with different types of prediction model in Table~\ref{tab:ab}.

\subsection{Experimental Analysis}
\label{sec:analysis}

\begin{table*}[]
\centering
\caption{The effects of different types of prediction models on the TextCNN decision model. 
The PRE columns are the BLEU score of prediction models generated queries or answers. \textit{WoA} (Wait-or-Answer) columns are decision model's~(DEC) accuracy score. 
}
\resizebox{1\textwidth}{!}{
\begin{tabular}{lcccccccccc}
\toprule
\multicolumn{2}{l}{\multirow{3}{*}{PRE. Type}} &
  \multicolumn{3}{c}{MultiWOZ} &
  \multicolumn{3}{c}{DailyDialog} &
  \multicolumn{3}{c}{CCPE} \\ \cline{3-11} 
\multicolumn{2}{l}{} &
  \multicolumn{2}{c}{PRE (BLEU)} &
  DEC (Acc.) &
  \multicolumn{2}{c}{PRE (BLEU)} &
  DEC (Acc.) &
  \multicolumn{2}{c}{PRE (BLEU)} &
  DEC (Acc.) \\ \cline{3-11} 
\multicolumn{2}{l}{}    & Agent & User           & WoA   & Agent & User           & WoA   & Agent & User          & WoA   \\ \hline
\multicolumn{2}{l}{N/A} & -     & -              & 77.68 & -     & -              & 75.79 & -     & -             & 68.65 \\ \hline
\multirow{2}{*}{LSTM} &
  Agent Prediction &
  \textbf{11.77} &
  0.80 &
  \multirow{2}{*}{80.04} &
  \textbf{4.51} &
  0.61 &
  \multirow{2}{*}{76.37} &
  \textbf{15.71} &
  0.00 &
  \multirow{2}{*}{70.04} \\
    & User Prediction   & 0.30  & \textbf{8.87}  &       & 0.15  & \textbf{8.70}  &       & 0.00  & \textbf{1.14} &       \\ \hline
\multirow{2}{*}{LSTM + Attn.} &
  Agent Prediction &
  \textbf{12.47} &
  0.72 &
  \multirow{2}{*}{\textbf{80.75}} &
  \textbf{19.19} &
  0.60 &
  \multirow{2}{*}{\textbf{79.02}} &
  \textbf{23.86} &
  0.00 &
  \multirow{2}{*}{\textbf{73.32}} \\
    & User Prediction   & 0.24  & \textbf{9.71}  &       & 0.26  & \textbf{24.52} &       & 0.00  & \textbf{1.46} &       \\ \hline
\multirow{2}{*}{LSTM w/ GLOVE + Attn.} &
  Agent Prediction &
  \textbf{13.37} &
  0.67 &
  \multirow{2}{*}{80.38} &
  \textbf{19.01} &
  0.67 &
  \multirow{2}{*}{78.56} &
  \textbf{19.56} &
  0.00 &
  \multirow{2}{*}{71.62} \\
    & User Prediction   & 0.51  & \textbf{10.61} &       & 0.21  & \textbf{24.65} &       & 0.00  & \textbf{1.77} &       \\ \bottomrule
\end{tabular}
}
\label{tab:ab}
\end{table*}
\noindent \textbf{Shortcomings of Tranditional Rule-based Models} \quad
As shown in Table \ref{tab:rea_results}, we can find out that though the ATLU and PTSU models are easy to implement, their results on all datasets are unsatisfactory and unstable. ATLU achieves 39.16 and 43.46 in terms of accuracy. PTSU has slightly better performance than ATLU. Both of two rule-based model have low performance in terms of F1 scores. This indicates that the rules based models have difficulty in this Wait-or-Answer task. The phenomenon that ATLU achieves 39.16 on E-COMM and 43.46 on After-Sales in terms of accuracy shows that the performance of rule-based models are unstable and highly rely on the habits of users in datasets. A significant difference between Wait-or-Answer task and incomplete utterance classification task is that all utterances in Wait-or-Answer task are complete. A long utterance does not mean the intention is complete and a short utterance does not represent an incomplete dialogue. Therefore, rule-based models rely on the surface features of utterances e.g., length of sentences and grammars, are not reliable on all situations in the Wait-or-Answer task.

\noindent \textbf{Benefits Brought By PTD Framework} \quad
From Table \ref{tab:rea_results}, we can see that our BERT-PTD model achieves the best performance on all datasets, e.g. BERT-PTD achieves 71.14 and 82.86 on After-Sales in terms of Accuracy and F1. Both are the highest performance on this dataset.
Besides, the other two PTD models: GRU-PTD and TextCNN-PTD also significantly outperform their corresponding baselines, e.g. BERT-PTD achieves 71.98 on E-COMM in terms of Accuracy, which is the lowest score among all PTD models but still outperforms all other baseline models (the highest baseline model Bi-GRU achieves 71.96). Even the most rudimentary PTD model: GRU-PTD can beat all baselines~(GRU, TextCNN, and BERT) in all these datasets. 

The results on more evaluation metrics on public datasets in Table~\ref{tab:f1} also verify that our PTD framework is a more suitable choice for the Wait-or-Answer task. In most cases, the PTD models have better precision and recall scores, this shows PTD framework will decrease the possibility of presenting false positive and false negative results. We can also find that sometimes baseline will get higher recall but lower precision and F1, e.g. BERT gets 89.46 on MultiWOZ and TextCNN gets 78.49 on DailyDialog in terms of recall. This shows that baseline models can make false positive decisions more frequently. After analyzing some results samples, e.g. as shown in Table \ref{tab:example}, we can find that because of the supplement of generative models, PTD models are better at understanding the intuitions of users behind dialogue context. In contrast, the baseline models may make wrong decisions when meeting a situation with complete semantic and syntax dialogue context. 

Above all, experimental results demonstrate that PTD model will improve the performance on the wait-or-answer task and decrease the possibility of making false positive and false negative decisions.

\noindent \textbf{PTD's Advantage on Small-scale Datasets} \quad
One of the most crucial limits of the dialogue systems' applications is the lack of high-quality datasets. In this case, we analyze the PTD's performance on small-scale datasets. 
CCPE is relatively small-scale datasets, which consists of only 502 dialogues and significantly more average turns (9.7 in train set compared with 4.09 in DailyDialog and 6.32 in MultiWOZ). Besides, the numbers of positive (Agent Answer) samples and negative (Agent Wait) samples are more imbalanced. This makes it much more difficult to train a satisfactory model.

As shown in Table \ref{tab:f1}, we can see that baselines: Bi-GRU, TextCNNs, and BERT achieve accuracy scores of 67.53, 68.65, and 70.99. We can observe that baselines' performance is significantly worse than that on large-scale datasets such as MultiWOZ. 
However, our PTD models: GRU-PTD, TextCNN-PTD, and BERT-PTD all achieve improved scores. 
As shown in Table \ref{tab:ab}, we can see that in small datasets, the prediction models' BLEU scores are not worse than that on larger datasets like MultiWOZ. And all type of prediction models help decision models get significant improvement, and improvement is positively correlated with the prediction models' performance. The LSTM with Attention-based prediction models gets the best generation scores and the best decision model results.
In this case, we can conclude that on small-scale and imbalanced datasets, baselines have more difficulty in achieving satisfying results. However, our PTD models can learn more semantic information from the dialogue history with the user prediction model and the agent prediction model. In this way, our PTD models can achieve much more satisfying results than the baselines. This improvement can be explained by the fact that the prediction model in PTD can exploit the information in dialogue history more thoroughly. 

\noindent \textbf{Effects of Prediction Models in PTD Framework}  \quad
Another interesting issue about the PTD framework is the prediction models' effects on the PTD framework. We investigate this issue by answering the following two questions: 

    \noindent(1). \textit{Do the prediction models work as we expect?}\quad We want to check out if the user prediction model can truly predict the user's supplementary input and the agent prediction model can predict the dialogue system's answer precisely. But, do they work as we expect? We conduct an experiment on the MultiWOZ dataset. As shown in Table~\ref{tab:ab}, the LSTM based agent prediction model get the BLEU score at 11.77 on agent samples, in which the ground-truth is agents' utterances, and the user prediction model gets the BLEU score at 0.3 on agent samples. Similar results are shown in other prediction models' experiments.
    This phenomenon doesn't mean that the user prediction model runs terrible. Actually, these results show that our user prediction model successfully behaves like a user. And its difficulty in generating agent utterance also meets our design. The example is also shown in Table \ref{tab:example}, in which the predicted agent utterance by user prediction model seems a high-quality fluent sentence and is also suitable for the scene. However, referring to the dialogue history, it is not a good choice since user in the last turn has said a sentence with similar intention \textit{what is the address for pizza hut in cherry hinton}, so the user prediction models' prediction \textit{i am looking for a guesthouse} is not a good choice for decision, which means, the decision model prefers to answer user's utterance. 
    From above we can conclude that contrasting results of the two prediction models work as we expect and help the decision model in \textit{Wait-or-Answer} task.
    
    \noindent(2). \textit{Can better prediction model lead to better PTD models?}\quad Another interesting question is that if the improvement of the prediction model can always boost the performance of PTD models. Take the DailyDialog as an example, we can see that with the enhancement of the attention mechanism and pre-trained GLOVE, the prediction models' performance increases~\footnote{With the attention mechanism, the BLEU score increases from 4.51 to 19.19. With the attention mechanism and pre-trained GLOVE vector, the BLEU score increases from 4.51 to 19.01.}. The accuracy of the PTD models also increases: from 76.37 to 79.02, from 76.37 to 78.56. We can also observe the same phenomenon on MultiWOZ. From those results, we can conclude that there is a positive correlation between the performance of prediction models and the final performance of PTD models. 
    From the above analysis, we can conclude that both the user prediction model and the agent prediction model can significantly enhance the decision models by predicting the dialogue interaction behavior. Recently, transformer based models show its effectiveness in generation tasks, e.g. BART~\cite{lewis2020bart}, T5~\cite{DBLP:journals/jmlr/RaffelSRLNMZLL20} and GPT~\cite{NEURIPS2020_1457c0d6}, they may further improve performance in PTD. We leave it as a future work.
    
\section{Related Work}
In this Section, we describe some work related to the wait-or-answer task, concentrating on response triggering recognizing and predicting incomplete user utterance.

Coman et al. \cite{DBLP:conf/interspeech/ComanYM0R19} implement an incremental Dialog State Tracker which is updated on a token basis to identify the point of maximal understanding in an ongoing utterance. DeVault
et al. \cite{DBLP:conf/sigdial/DeVaultST09} propose a method for determining when a system has reached a point of maximal understanding of an ongoing user utterance to responsive overlap behaviors in dialogue
systems, opening possibilities for systems
to interrupt, acknowledge or complete a
user’s utterance while it is still in progress. Liu et al. \cite{DBLP:conf/kdd/LiuJXYY20} highlight the issues in previous models that in a supervised setting, the response timing information in the dialogues may be inaccurate in real life scenarios, such as customer service. MRTM address the inappropriate triggered responses problem very similar to wait-or-answer situation. MRTM is a self-supervised learning scheme leveraging the semantic matching relationships between the context and the response to train a semantic matching model and obtains the weights of the co-occurring utterances in the context through an asymmetrical self-attention mechanism. PTD adopts this similar settings that all training labels, e.g. turn tags and role tags are generated from dialogue history without human intervention. MRTM mainly concentrates on response selection models, while PTD explores to solve this task in a more general 
application scenario.

Some researches \cite{DBLP:journals/corr/abs-2008-01474} have also investigated the incomplete utterance restoration and rewriting. Liu et al. \cite{DBLP:conf/emnlp/LiuCLZZ20} formulate the incomplete utterance rewriting as a semantic segmentation task and propose a model for predicting the edit operations in parallel. Pan et al. \cite{DBLP:conf/emnlp/PanBWZL19} facilitate the study of incomplete utterance restoration for open-domain dialogue systems and propose a “pick-and-combine” model to restore the incomplete utterance from its context.

\section{Conclusion} \label{sec:conclusion}
Conventional dialogue systems require that users must describe their intents in a single utterance, otherwise dialogue systems will answer immediately, which may cause misunderstanding or reply to the wrong question.
In this paper, we explicitly define the aforementioned quandary as a novel Wait-or-Answer task. We further propose a framework named Predict-then-Decide~(PTD) model to tackle with this Wait-or-Answer task. 
Our paper sheds light on the enhancement of the existing dialogue systems' ability to handle the wait-or-answer problem. 
We believe that the clearly-defined Wait-or-Answer task and PTD framework can provide an interesting topic for both academic and industrial NLP communities. 
\iffalse
Conventional dialogue systems require that users must describe their intents in a single utterance, otherwise dialogue systems will answer immediately, which may cause misunderstanding or reply to the wrong question. Motivated by this problem, we explicitly define a novel task named Wait-or-Answer
and propose an Predict-then-Decide~(PTD) model to tackle with this Wait-or-Answer task. 
Experimental results demonstrate that our PTD models achieve a great extent of improvement over baselines on addressing this Wait-or-Answer task. 
\fi

\section*{Acknowledgments}
This work was supported by the National Natural Science Foundation of China (No. 62072399, No. 61402403, No. U19B2042), MoE Engineering Research Center of Digital Library, Chinese Knowledge Center for Engineering Sciences and Technology, Alibaba-Zhejiang University Joint Research Institute of Frontier Technologies, and the Fundamental Research Funds for the Central Universities.
\bibliographystyle{IEEEtran.bst}
\bibliography{sample}
\begin{IEEEbiographynophoto}{Zehao Lin} received his B.Sc. degree from Zhejiang University in 2015. He is currently working toward the Ph.D. degree in the College of Computer Science and Technology, Zhejiang University. His current research interests include natural Language processing, dialogue systems, and multimodal.
\end{IEEEbiographynophoto}

\begin{IEEEbiographynophoto}{Shaobo Cui} is a senior algorithm engineer in DAMO Academy, Alibaba Group. His current research interests include machine learning, mathematical optimization, natural language processing, and their applications. Previously, he received his M.Sc. degree from Tsinghua University.
\end{IEEEbiographynophoto}

\begin{IEEEbiographynophoto}{Guodun Li} received his B.Sc. degree from Hangzhou Dianzi University in 2020. He is currently working toward the M.Sc. degree in the College of Computer Science and Technology, Zhejiang University. His current research interests include natural language processing and their applications.
\end{IEEEbiographynophoto}

\begin{IEEEbiographynophoto}{Xiaoming Kang} received his M.Sc. degree from Fudan University in 2016. He is now a Senior Algorithm Engineer at Alibaba Group. His current research interests include natural language processing algorithms and their applications. He mainly focused on Dialogue System and Question answering.
\end{IEEEbiographynophoto}

\begin{IEEEbiographynophoto}{Feng Ji} received his B.Sc. degree from Tongji University and Ph.D. degree from Fudan University in 2003 and 2012 respectively. He was a senior algorithm engineer in Alibaba Group and now currently works in Tencent. His research interests include artificial intelligence, natural language understanding and generation, dialogue systems, information retrieval and recommendation systems.
\end{IEEEbiographynophoto}

\begin{IEEEbiographynophoto}{Feng-Lin Li} received his Ph.D. degree from University of Trento in 2016. He is currently working at DAMO Academy, Alibaba Group. His research interests include natural language processing and knowledge graph.
\end{IEEEbiographynophoto}

\begin{IEEEbiographynophoto}{Zhongzhou Zhao} received his M.Sc. degree in computer science from the Harbin Institute of Technology and the University of Pavia. He is currently a staff algorithm engineer at DAMO Academy, Alibaba Group. He is one of the early members of AliMe chatbot, and is currently responsible for the AI algorithm aspect of AliMe Avatar. His research interests include MRC, NLG, VQA and Multimodal Understanding. 
\end{IEEEbiographynophoto}

\begin{IEEEbiographynophoto}{Haiqing Chen} received his B.S degree from Zhejiang University of technology in 2009. Now He is a senior algorithm expert of Alibaba cloud intelligence business group Damo Academic. His current research interests include natural language processing, question \& answering, dialogue, machine learning and deep learning.
\end{IEEEbiographynophoto}

\begin{IEEEbiographynophoto}{Yin Zhang} received his Ph.D. degree from Zhejiang University in 2009. He is currently an associate professor with the College of Computer Science and Technology, Zhejiang University, China. His research interests include machine reading comprehension, question answering systems, multi-agent systems, and digital library.
\end{IEEEbiographynophoto}

\end{document}